\documentclass[a4paper,11pt]{article}
\pdfoutput=1 
\usepackage[dvipsnames]{xcolor}
\usepackage{float}
\usepackage{amssymb}
\usepackage{jinstpub} 
\usepackage{booktabs}
\usepackage{array}
\usepackage{multirow}
\usepackage{hyperref} 
\usepackage{graphicx} 
\usepackage{url}  

\title{Benchmarking Vision-Language Models on Optical Character Recognition in Dynamic Video Environments}

\author{Sankalp Nagaonkar, Augustya Sharma, Ashish Choithani, Ashutosh Trivedi\footnote{Corresponding Author}}

\emailAdd{engg@videodb.io}

\abstract{
This paper presents an open-source initiative by VideoDB~\cite{videodb} to benchmark Vision-Language Models (VLMs) on Optical Character Recognition (OCR) tasks in dynamic video environments. To support this effort, we introduce a meticulously curated video dataset containing 1,477 manually annotated frames across diverse domains, such as code editors, news broadcasts, YouTube videos, and advertisements. We benchmarked three leading Vision-Language Models (Anthropic Claude-3 \cite{anthropic2024claude}, Gemini-1.5 \cite{google2024gemini}, and OpenAI GPT-4o \cite{openai2024gpt4}) alongside traditional Computer Vision (CV) OCR systems (EasyOCR \cite{EasyOCR2024} and RapidOCR \cite{RapidOCR2021}). Performance metrics such as Word Error Rate (WER), Character Error Rate (CER), and Accuracy were used to evaluate and compare these models. This study provides valuable insights into the capabilities and limitations of these models in real-world video OCR tasks. The dataset and benchmarking code are publicly available on GitHub at \url{https://github.com/video-db/ocr-benchmark}.
}

\keywords{Optical Character Recognition (OCR), Benchmark, Dataset, Vision-Language Models (VLMs), Video Processing, VideoDB}

\begin{document}
\maketitle
\flushbottom

\section{Introduction}
\label{sec:intro}

Optical Character Recognition (OCR) is a foundational technology in artificial intelligence, enabling the extraction of textual information from visual content. With the advent of Vision-Language Models (VLMs), there is a growing interest in exploring their potential to outperform traditional OCR methods, particularly in dynamic video environments. However, an important question arises: Can VLMs fully replace domain-specific OCR systems?

\noindent To answer this, we present a comprehensive benchmarking study using a newly developed dataset comprised of manually annotated frames spanning a variety of domains, including code editors, news channels, YouTube videos, advertisements, online lectures, and more. This work provides an in-depth evaluation of both VLMs and established Computer Vision-based OCR techniques under video-based settings. Our key contributions are as follows.
\begin{itemize}
    \item \textbf{Dataset Introduction}: We introduce a novel dataset that contains 1,477 annotated video frames covering various real-world domains such as code editing tools, news broadcasts, YouTube channels, advertisements, and online lectures.
    
    \item \textbf{Comprehensive Benchmarking}: We evaluated and compared the performance of state-of-the-art Vision-Language Models (\textit{e.g.}, Claude-3 \cite{anthropic2024claude}, Gemini-1.5 \cite{google2024gemini}, and GPT-4o \cite{openai2024gpt4}) and traditional Computer Vision OCR systems (\textit{e.g.}, EasyOCR \cite{EasyOCR2024}, RapidOCR \cite{RapidOCR2021}) using metrics such as Word Error Rate (WER), Character Error Rate (CER) and overall accuracy.
    \item \textbf{Open-Source Contributions}: To encourage further research, we publicly release the data set and benchmarking process under \href{https://github.com/video-db/ocr-benchmark/blob/main/LICENSE}{MIT License} via VideoDB~\cite{videodb}, allowing researchers to easily evaluate new models.
\end{itemize}

\noindent This paper is structured as follows: Section~\ref{sec:related_work} reviews key related work in the domains of Optical Character Recognition (OCR) using the traditional approach, and Sections ~\ref{review_sota} outlines a basic overview of Vision-Language Models (VLMs), providing context for our contributions. Section~\ref{sec:data_desc} describes the meticulous process of creating and curating our dataset, highlighting its diversity and relevance to real-world applications. Section~\ref{sec:methodology} outlines the benchmarking methodology, detailing the evaluation metrics and experimental setup used to compare state-of-the-art models. Section~\ref{sec:results} presents the evaluation results, accompanied by comprehensive benchmarking charts and visualizations for in-depth analysis. Lastly, Section~\ref{sec:conclusion} summarizes our findings, discusses the broader implications of the results, and outlines potential directions for future research.

\noindent Additionally, we invite readers to explore samples from our dataset along with their corresponding ground truth annotations, which are included as supplementary material in Section~\ref{sec:supplementary_material} to provide deeper insights into the dataset structure and quality.

\section{Related Work}
\label{sec:related_work}

Several open-source Optical Character Recognition (OCR) frameworks have been proposed, each tailored to meet different performance and usability requirements.\\ 

\subsection{Traditional Approach}
\noindent RapidOCR \cite{RapidOCR2021} stands out as a high-performance OCR framework that uses ONNXRuntime, OpenVINO, and PaddlePaddle \cite{paddleocr2023} to provide fast inference across platforms, including servers, mobile devices, and embedded systems. It supports multilingual OCR tasks and provides pre-trained models, making it ideal for real-time applications that require high throughput.

\noindent EasyOCR \cite{EasyOCR2024} is another lightweight OCR toolkit that employs a two-stage approach: text detection using the CRAFT algorithm (Character Region Awareness for Text Detection) \cite{baek2019character}, followed by text recognition using a Convolutional Recurrent Neural Network (CRNN) with a Connectionist Temporal Classification (CTC) \cite{graves2012connectionist} decoder.

\section{Vision-Language Models}
\label{review_sota}
Vision-Language Models (VLMs) have made remarkable advancements, positioning themselves as potential universal solutions for a wide range of tasks that traditionally required separate models for vision and language processing.  By integrating state-of-the-art advancements in computer vision and natural language processing, Vision-Language Models (VLMs) are enabling a wide range of multimodal tasks. Their generalizability suggests they may replace dedicated, task-specific architectures. This transformative capability underscores the importance of analyzing their performance across diverse applications, including Optical Character Recognition (OCR), as explored in this paper. Understanding their strengths and limitations in such specialized tasks is crucial for assessing their readiness to become the go-to models for every domain.

\subsection{Overview of VLMs}
VLMs learn joint representations of images and text through multimodal architectures, typically combining vision encoders with large language models. Training on extensive multimodal datasets allows these models to grasp complex visual semantics and contextual language usage.

\subsection{State-of-the-Art Models}
The system integrates multiple VLMs to compare their performance.
\begin{itemize}
        \item \texttt{Anthropic:} Claude-3 Sonnet \cite{anthropic2024claude} improves its predecessors by focusing on intelligence and speed. It integrates a robust visual encoder with a large language decoder, excelling in tasks like VQA and multimodal reasoning \cite{lee2024vhelm}. Benchmarks indicate that Claude-3 Sonnet outperforms competitor models and previous versions in various evaluations, demonstrating superior reasoning and content generation capabilities. \\
        \item \texttt{Google:} Gemini-1.5 Pro \\ Gemini-1.5 Pro \cite{google2024gemini} combines advanced visual feature extraction with text generation within a multimodal transformer architecture. Its extensive pretraining on video-text datasets positions it as a leading model for video understanding tasks. Gemini-1.5 Pro exhibits strong performance in benchmarks such as MSR-VTT \cite{xu2016msr}, \cite{de2025video} and TVQA \cite{lei2018tvqa}, reflecting its proficiency in video-related subtasks.\\
        \item \texttt{OPENAI:} GPT-4o \\ GPT-4o is an evolution of the GPT-4 \cite{openai2024gpt4} architecture, featuring an expanded context window and enhanced processing speed. While primarily a language model, its capabilities extend to multimodal tasks. It is the most advanced variant of the GPT series, and it has achieved superior performance on various multimodal benchmarks.

        \end{itemize}



\section{Data Description and Processing}
\label{sec:data_desc}
We propose an efficient approach to create vision-language model (VLM) datasets from videos using \textbf{VideoDB} \cite{videodb}. With its image extraction algorithms and indexing capabilities, VideoDB automates the process of extracting and organizing images, eliminating the need for manual collection and management. This streamlined workflow simplifies the creation of data sets, enabling scalable and efficient processing of visual and textual data from videos.

\noindent We created a custom data set of 1,477 frames across various domains, Code editors, News channels, YouTube videos, Advertisements, Talk shows, Online lectures, Traffic rules, and more. Examples of the data set are provided in the supplementary material. (see Section~\ref{dataset_examples})

\section{Evaluation Metrics Calculation}
\label{sec:methodology}
The key metrics include:
\begin{itemize}
    \item \textbf{Character Error Rate (CER):} Measures the edit distance between the ground-truth text and the OCR output at the character level. It's calculated as:
    
    \begin{equation}
    \text{CER} = \frac{(S + D + I)}{N}
\end{equation}

    where $S$ is the number of substitutions, $D$ is the number of deletions, $I$ is the number of insertions, and $N$ is the total number of characters in the ground truth. (Lower is better)
    \item \textbf{Word Error Rate (WER):} Similar to CER, but at the word level. (Lower is better)
    \item \textbf{Accuracy:} Calculated as:
    \begin{equation}
        (1 - CER) \times 100
    \end{equation}
    providing a percentage measure of how accurate the OCR output is compared to the ground truth. (Higher is better)
\end{itemize}

\section{Results}
\label{sec:results}
\subsection{Qualitative Results}

In this section, we present a detailed comparison of ground truth with model outputs, analyzing differences in sentence structure, content preservation, and clarity. This analysis includes identifying character-level additions, substitutions, and omissions compared to the ground truth.






\begin{figure}[htbp]
\centering

\begin{minipage}{\textwidth}
    \centering
    \includegraphics[width=0.9\linewidth]{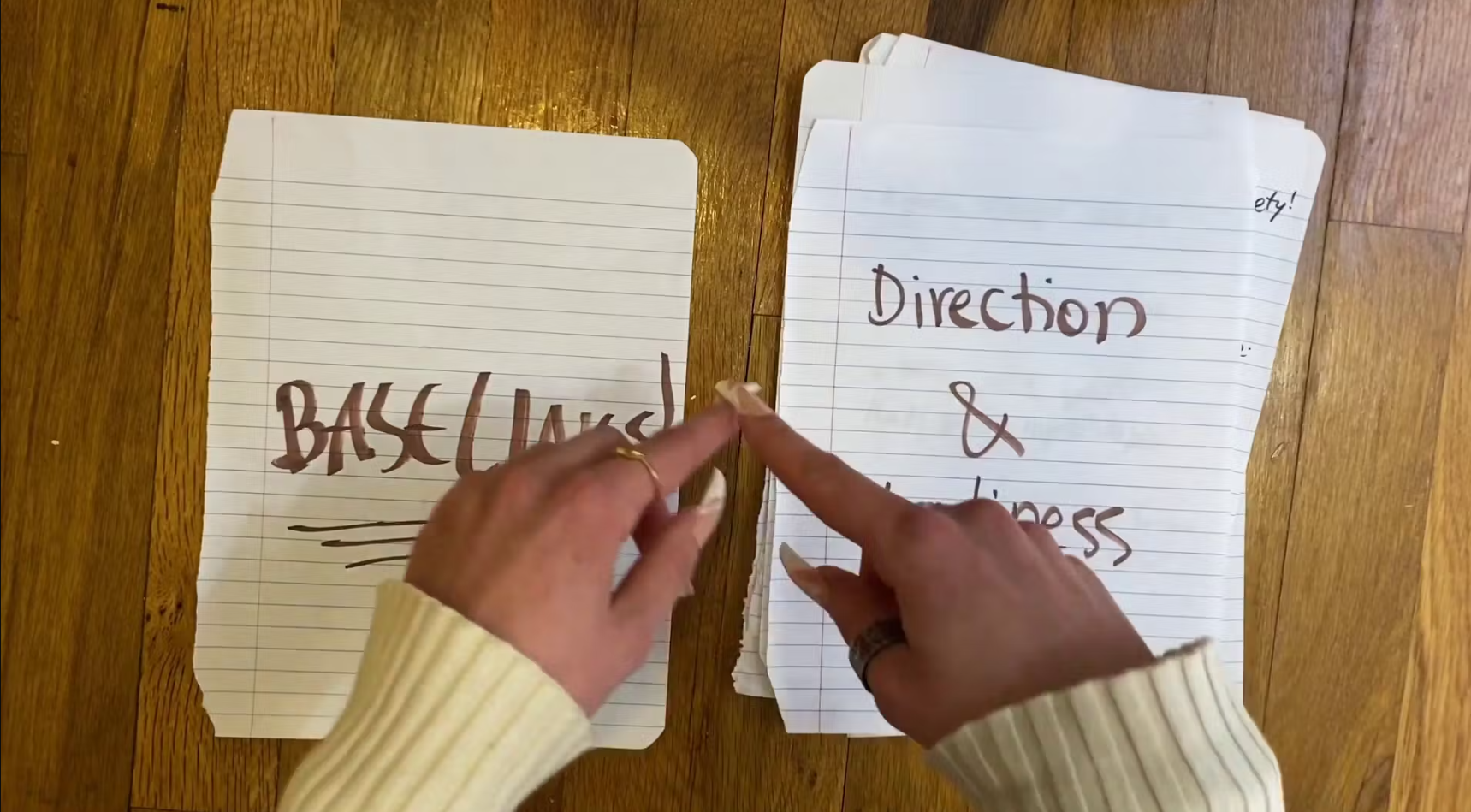}
\end{minipage}

\vspace{2.5em}

\renewcommand{\arraystretch}{1.3}
\begin{tabular}{|
    >{\centering\arraybackslash}p{0.18\textwidth} 
    |>{\centering\arraybackslash}p{0.27\textwidth} 
    | 
    >{\centering\arraybackslash}p{0.18\textwidth} 
    |>{\centering\arraybackslash}p{0.27\textwidth}
    |}
    \hline
    \multicolumn{2}{|c|}{\large\textbf{Ground Truth}} & 
    \multicolumn{2}{c|}{\textcolor{ForestGreen}{\textit{BASE  Direction \& ss ety!}}} \\
    \hline\hline
    \multicolumn{2}{|c|}{\large\textbf{Vision-Language Models}} & 
    \multicolumn{2}{c|}{\large\textbf{Computer Vision Models}} \\
    \hline
    \multicolumn{1}{|c|}{\textit{Claude-3 Sonnet}} & \multicolumn{1}{c|}{\textit{ety! Direction \& \textcolor{red}{progr}ess Base\textcolor{red}{line}}} &
    \multicolumn{1}{c|}{\textit{RapidOCR}} & \multicolumn{1}{c|}{\textit{\textcolor{red}{ha} Direc\textcolor{red}{h}on BAE \textcolor{red}{n}ess}} \\
    \cline{1-2}\cline{3-4}
    \multicolumn{1}{|c|}{\textit{Gemini-1.5 Pro}} & \multicolumn{1}{c|}{\textit{BASE\textcolor{red}{LINE} Direction \& \textcolor{red}{n}ess ety!}} &
    \multicolumn{1}{c|}{\textit{EasyOCR}} & \multicolumn{1}{c|}{\textit{ety Di\textcolor{red}{v}ec\textcolor{red}{h}on Ba\textcolor{red}{k} \textcolor{red}{6Lt}}} \\
    \cline{1-2}
    \multicolumn{1}{|c|}{\textit{GPT-4o}} & \multicolumn{1}{c|}{\textit{BASE \textcolor{red}{Uses} Direction \& \textcolor{red}{Fitn}ess}} & & \\
    \hline
\end{tabular}

\caption{Handwritten and Occlulded text (Additons and Substitutions are marked in \textcolor{red}{Red})}
\label{fig:ocr_6}
\end{figure}

\noindent 
In figure \ref{fig:ocr_6}, All models encounter difficulty interpreting the text, particularly "ss ety." Claude misinterprets "BASE" as "Baseline" and introduces the term "progress". Gemini captures the phrase "Direction \&" but misreads "ss ety!" as "ness ety!" and substitutes "BASE" with "BASELINE." GPT-4 comes closer to the ground truth but misinterprets "ss ety!" as "Fitness" and substitutes "BASE" with "BASE Uses." The traditional computer vision models, however, demonstrated significant shortcomings, failing to recognize even simple text like "Direction". Furthermore, the models introduce spurious characters and omissions. RapidOCR adds an "n" to "BASE," rendering it "BAEness," while EasyOCR substitutes characters, incorrectly producing "BaK 6Lt." This highlights their limited ability to handle noisy or partially obscured characters compared to the Vision-Language Models, which, while imperfect, still capture more of the overall textual context.  This demonstrates the challenges these models face with occluded texts.

\pagebreak

\begin{figure}[htbp]
\centering

\begin{minipage}{\textwidth}
    \centering
    \includegraphics[width=0.7\linewidth]{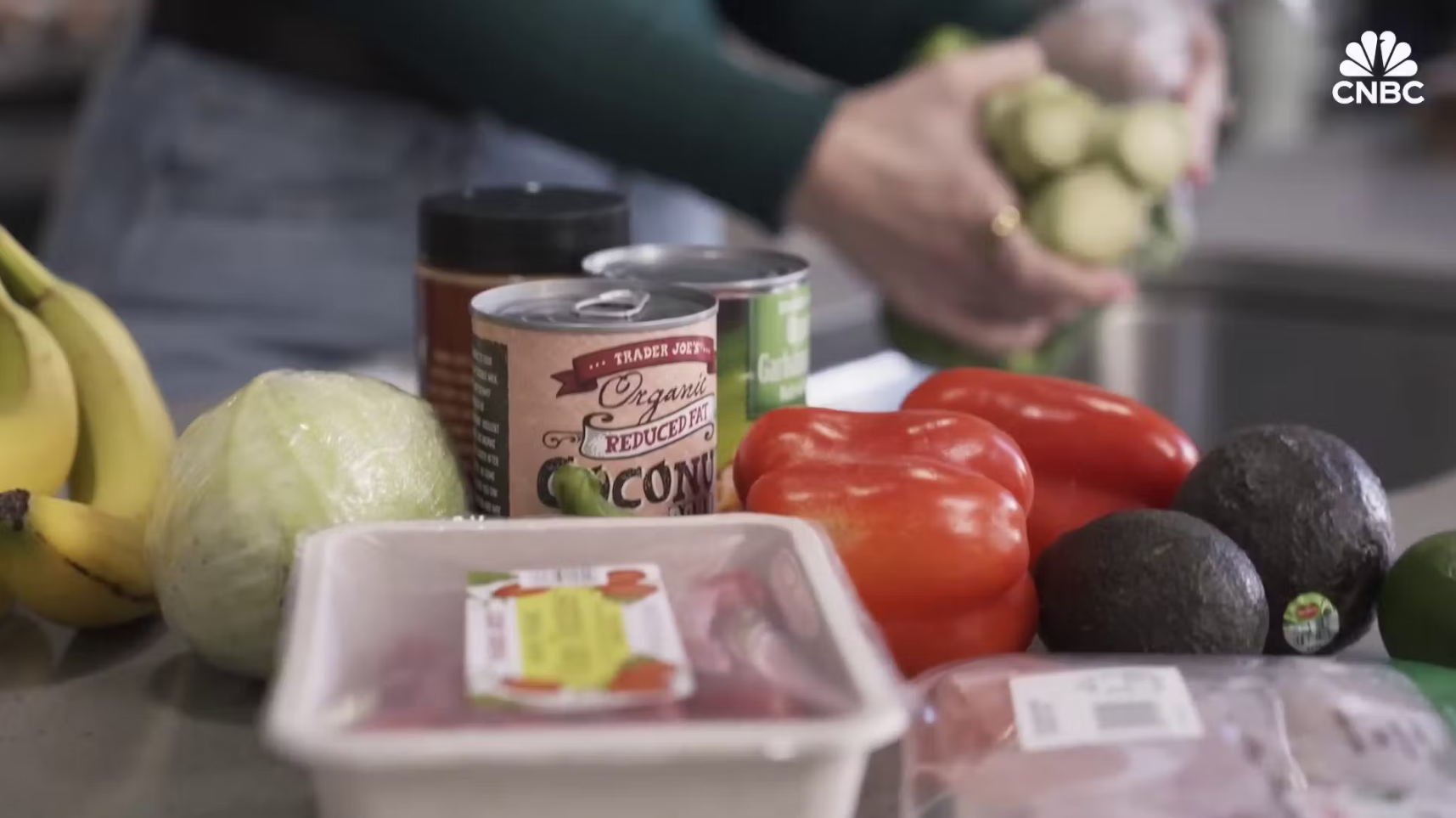}
\end{minipage}

\vspace{2.5em}

\renewcommand{\arraystretch}{1.6}  
\begin{tabular}{|
    >{\centering\arraybackslash}p{0.18\textwidth} 
    |>{\centering\arraybackslash}p{0.27\textwidth} 
    | 
    >{\centering\arraybackslash}p{0.18\textwidth} 
    |>{\centering\arraybackslash}p{0.27\textwidth}
    |}
    \hline
    \multicolumn{2}{|c|}{\large\textbf{Ground Truth}} & 
    \multicolumn{2}{|c|}{\parbox[c]{0.45\textwidth}{\vspace{0.5em}\centering\textcolor{ForestGreen}{\textit{"TRADER JOE'S\\Organic REDUCED FAT\\C CONU CNBC"}}\vspace{0.5em}}} \\
    \hline\hline
    \multicolumn{2}{|c|}{\large\textbf{Vision-Language Models}} & 
    \multicolumn{2}{c|}{\large\textbf{Computer Vision Models}} \\
    \hline
    \multicolumn{1}{|c|}{\textit{Claude-3 Sonnet}} & \parbox[c]{0.27\textwidth}{\vspace{0.5em}\centering\textit{Trader Joe's Organic Reduced Fat Co\textcolor{red}{co}nu\textcolor{red}{t} \textcolor{red}{Milk} CNBC}\vspace{0.5em}} &
    \multicolumn{1}{c|}{\textit{RapidOCR}} & \parbox[c]{0.27\textwidth}{\vspace{0.5em}\centering\textit{CNBC TRADER JOE'S. rgani REDUCEDFAT \textcolor{red}{e} CONU}\vspace{0.5em}} \\
    \cline{1-2}\cline{3-4}
    \multicolumn{1}{|c|}{\textit{Gemini-1.5 Pro}} & \parbox[c]{0.27\textwidth}{\vspace{0.5em}\centering\textit{TRADER JOE'S Organic REDUCED FAT CONU CNBC}\vspace{0.5em}} &
    \multicolumn{1}{c|}{\textit{EasyOCR}} & \parbox[c]{0.27\textwidth}{\vspace{0.5em}\centering\textit{CNBC \textcolor{red}{Jorte @xgad pb TK}ADER REDUCED\textcolor{red}{I} CON\textcolor{red}{I}}\vspace{0.5em}} \\
    \cline{1-2}
    \multicolumn{1}{|c|}{\textit{GPT-4o}} & \parbox[c]{0.27\textwidth}{\vspace{0.5em}\centering\textit{TRADER JOE'S Organic REDUCED FAT CO\textcolor{red}{CO}NU\textcolor{red}{T} CNBC}\vspace{0.5em}} & & \\
    \hline
\end{tabular}

\caption{TV Commercial (Additions and Substitutions are marked in \textcolor{red}{Red})}
\label{fig:ocr_5}
\end{figure}

\noindent In figure \ref{fig:ocr_5}, Claude's output introduces "Coconut Milk," which is not present in the ground truth, making it partially incorrect despite capturing the overall format and context. Gemini retains the truncated "CONU" from the ground truth and preserves the structure. GPT-4 provides the full product name by replacing "C CONU" with "COCONUT," which diverges from the ground truth's truncation, demonstrating an over-correction. While each model demonstrates partial success, none perfectly matches the ground truth. Claude and GPT-4 introduces extraneous content, and Gemini maintains the truncation but deletes a character. In contrast, the computer vision models fail to maintain proper capitalization and spacing. RapidOCR produces usable text, but EasyOCR performs significantly worse, outputting random, meaningless text. Additional results are provided in the supplementary material. (see Section~\ref{additional_results})

\subsection{Benchmark Results}
\begin{figure}[htbp]
    \begin{minipage}[b]{0.48\textwidth}
        \centering
        \includegraphics[width=\textwidth]{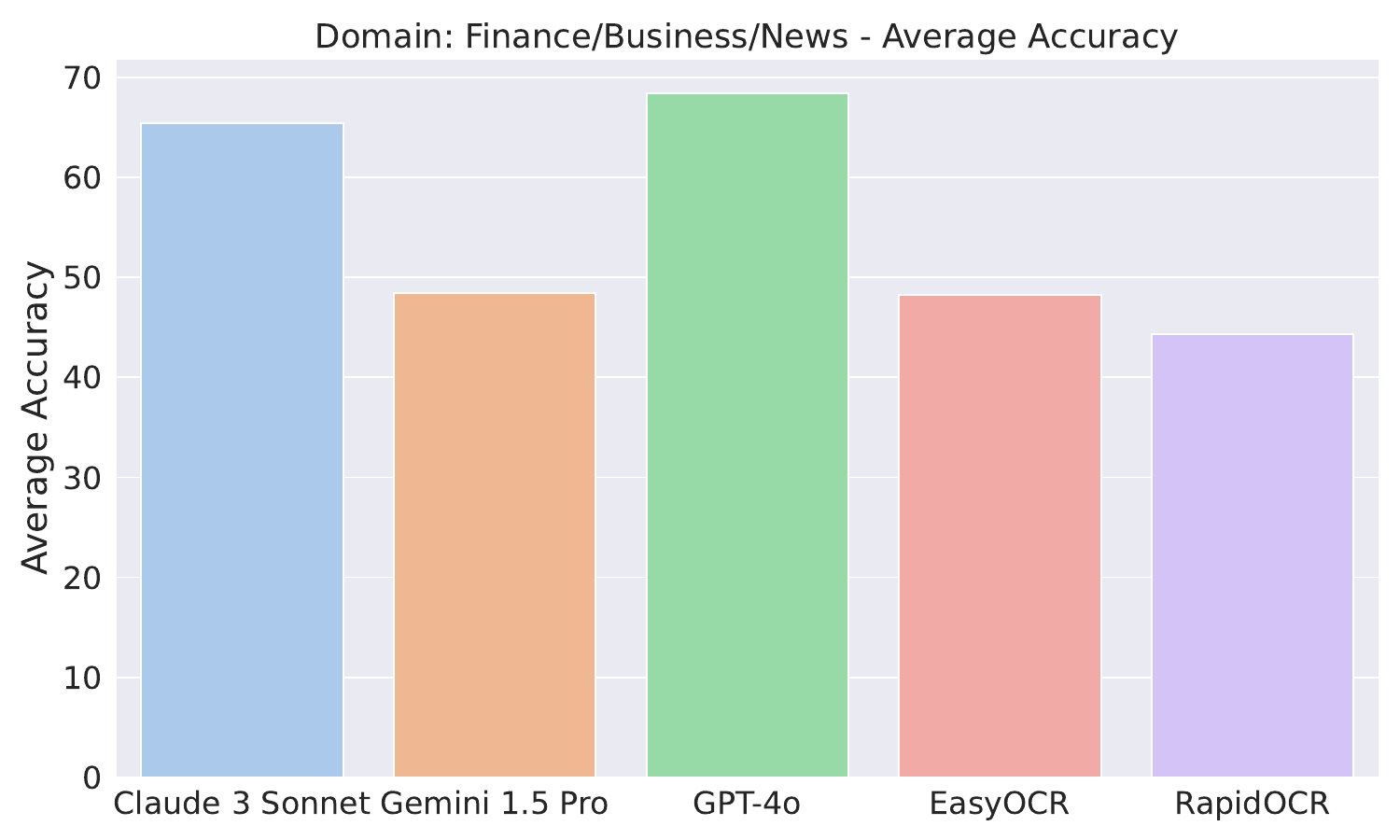}
        \caption{Finance/Business/News Text}
        \label{fig:domain_finance}
    \end{minipage}
    \hfill
    \begin{minipage}[b]{0.48\textwidth}
        \centering
        \includegraphics[width=\textwidth]{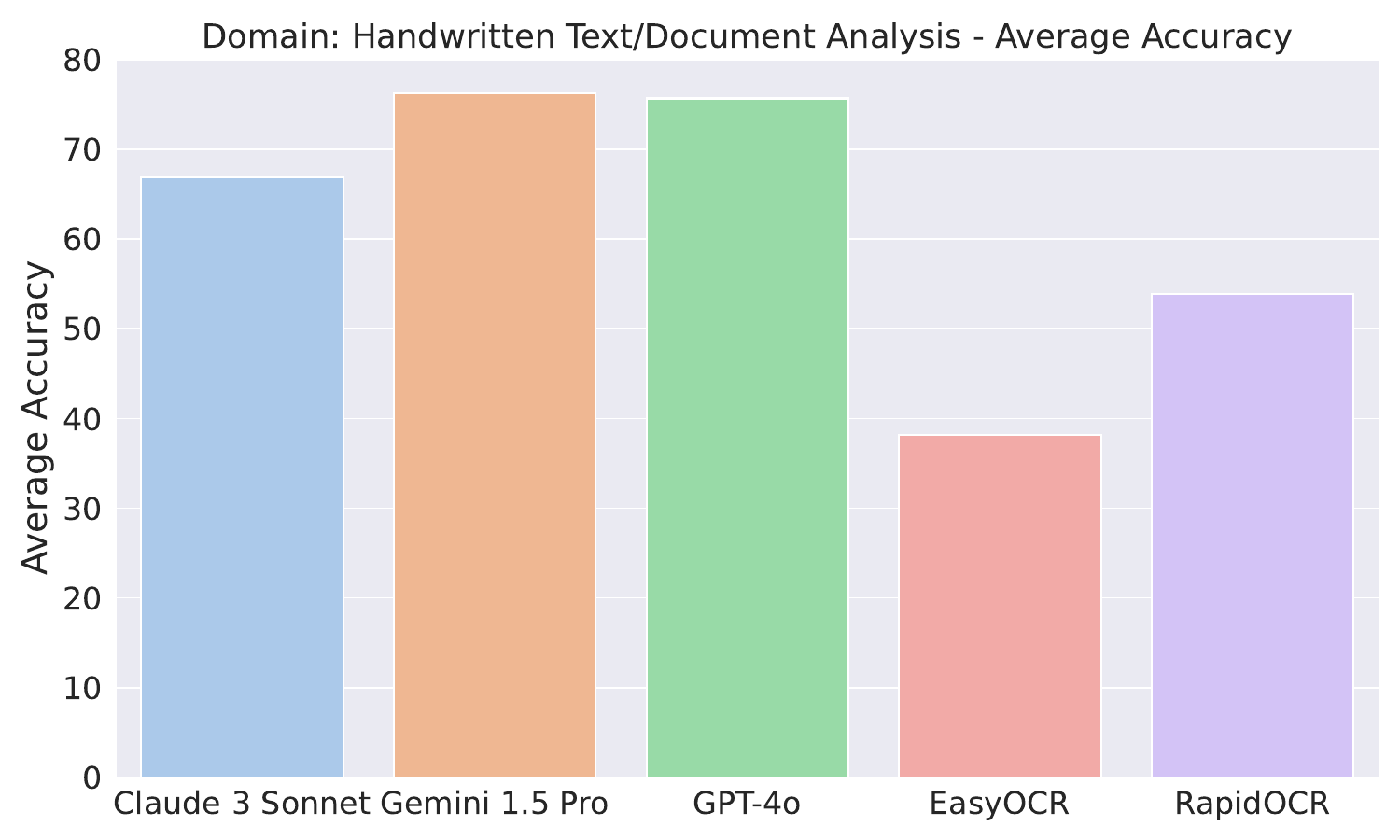}
        \caption{Handwritten Text}
        \label{fig:domain_handwritten}
    \end{minipage}
\end{figure}

\begin{figure}[htbp]
   \begin{minipage}[b]{0.48\textwidth}
       \centering
       \includegraphics[width=\textwidth]{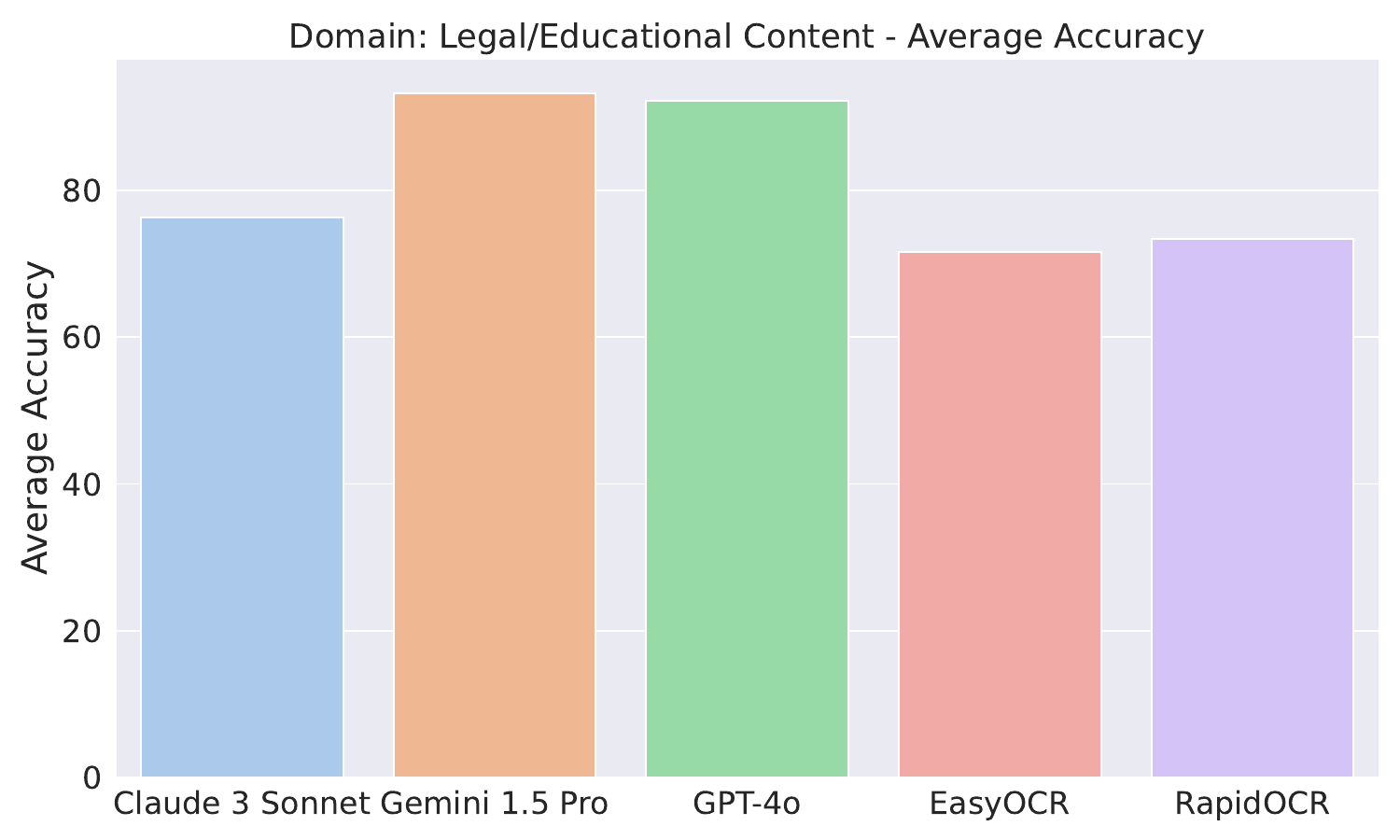}
       \caption{Legal/Educational Text}
       \label{fig:domain_legal}
   \end{minipage}
   \hfill
   \begin{minipage}[b]{0.48\textwidth}
       \centering
       \includegraphics[width=\textwidth]{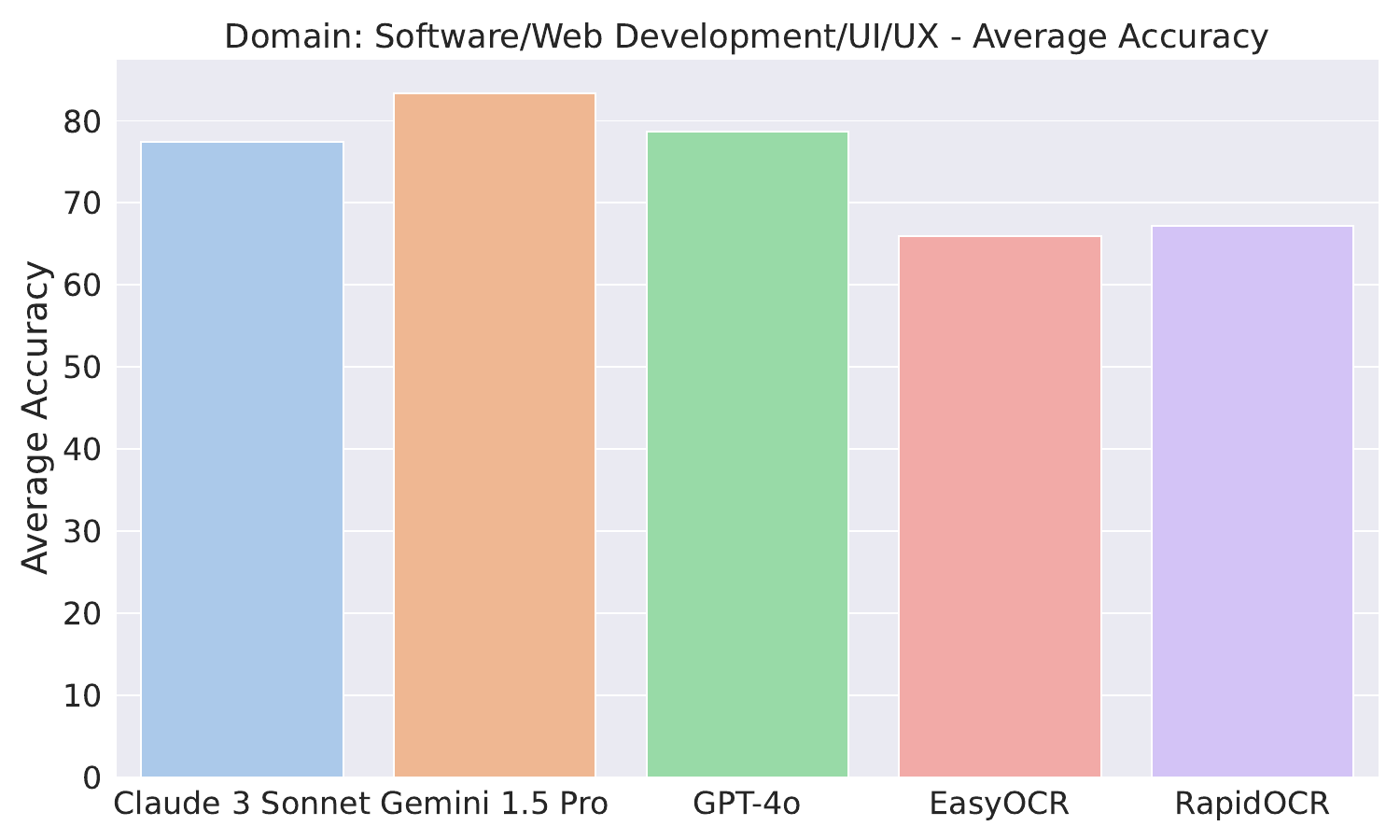}
       \caption{Software/Web Development/UI/UX Text}
       \label{fig:domain_software}
   \end{minipage}
\end{figure}

\begin{figure}[htbp]
    \centering
    \includegraphics[width=0.8\textwidth]{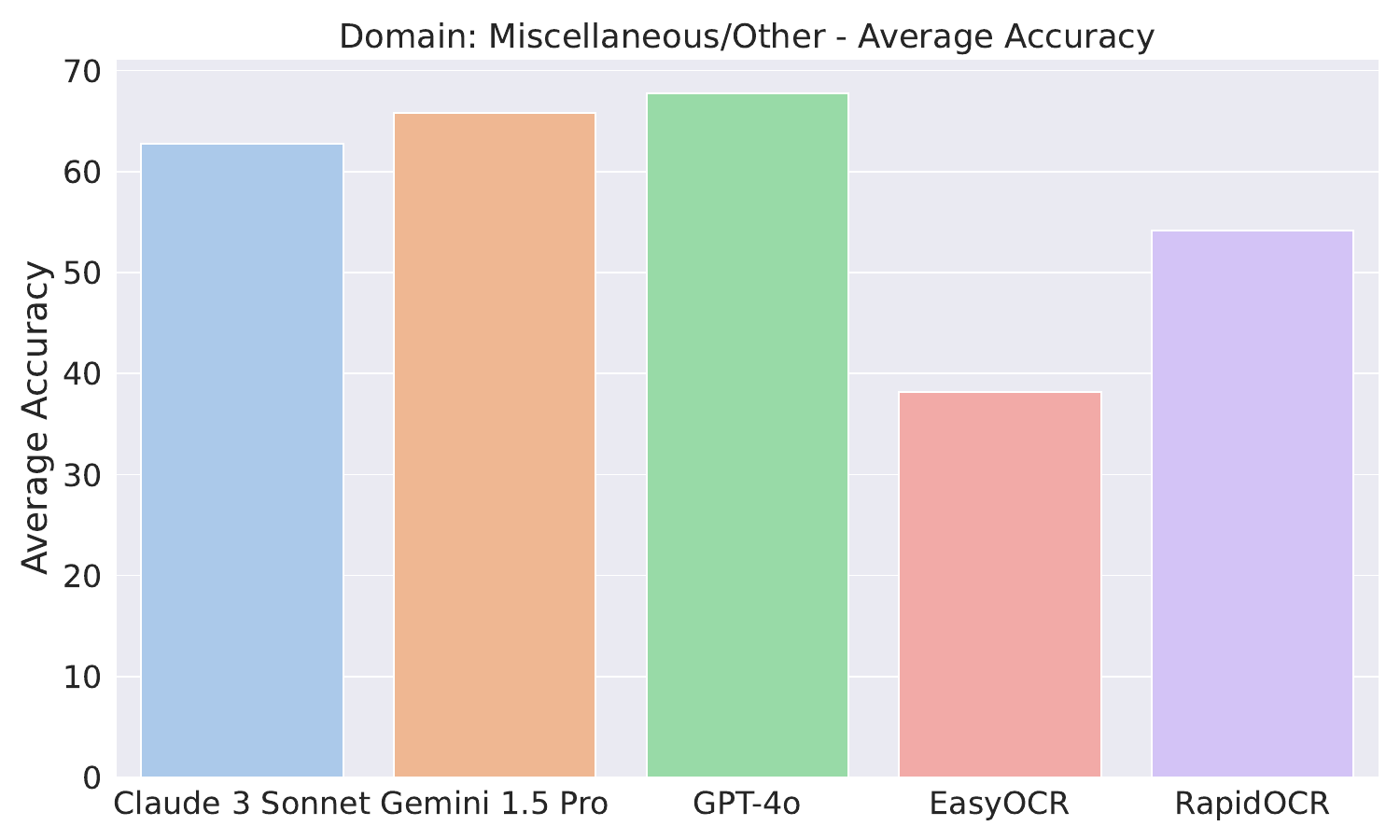}
    \caption{Miscellaneous/Other Text}
    \label{fig:domain_others}
\end{figure}

\begin{table}[H]
    \centering
    \scalebox{0.90}{
    \begin{tabular}{|l|c|c|c|c|}
\hline
\textbf{Model}  & \textbf{Character Error Rate (CER)} & \textbf{Word Error Rate (WER)} & \textbf{Average Accuracy (\%)} \\ \hline
RapidOCR           & 0.4302 & 0.7620 & 56.98 (\textcolor{red}{↓19.24}) \\ \hline
EasyOCR           & 0.5070 & 0.8262 & 49.30 (\textcolor{red}{↓26.92}) \\ \hline
Claude-3 Sonnet          & 0.3229 & 0.4663 & 67.71 (\textcolor{red}{↓8.51}) \\ \hline
Gemini-1.5 Pro          & 0.2387 & 0.2385 & 76.13 (\textcolor{red}{↓0.09}) \\ \hline
GPT-4o           & 0.2378 & 0.5117 & \textbf{76.22} \\ \hline
\end{tabular}}
\caption{Performance metrics of Vision-Language and Traditional Computer Vision Models}
\label{VLM_performance}
\end{table}
\noindent As shown in Table \ref{VLM_performance}, GPT-4o achieves the highest overall accuracy, while Gemini-1.5 Pro demonstrates the lowest word error rate. RapidOCR and EasyOCR perform poorly, with considerably higher error rates and lower accuracy. Claude-3 Sonnet's performance falls between the other models. In terms of processing time per image, GPT-4 was the slowest, followed by Claude, with Gemini demonstrating the fastest processing time. Furthermore, the graphs in Figures \ref{fig:domain_finance}, \ref{fig:domain_handwritten}, \ref{fig:domain_legal}, \ref{fig:domain_software}, and \ref{fig:domain_others} provide a comprehensive visualization of the average domain-wise precision achieved by different OCR and Vision-Language Models. GPT-4o demonstrates exceptional performance across all domains, consistently achieving accuracy rates between 65-80\%. In particular, it excels in legal / educational content with approximately 84\% accuracy, while maintaining robust performance in challenging domains like handwritten text. In contrast, Gemini-1.5 Pro shows significant performance variability, particularly struggling with Finance/Business/News content where its accuracy drops to around 50\%. Traditional OCR solutions like EasyOCR and RapidOCR consistently underperform compared to modern vision-language models.

\pagebreak

\noindent However, a significant downside of using VLMs for OCR tasks, and indeed for any task, lies in their content security policies. If the input content triggers these security flags, the model might refuse to generate any output. This can happen even if the content is benign, due to mistakenly triggered security protocols, and this impacts the reliability of these models in real-world applications. This means that while performance metrics like accuracy and speed are important, the dependability of VLMs is also contingent on their security systems and how prone they are to false positives. This factor needs to be carefully considered when choosing a VLM for practical deployment.

\section{Conclusion and Future Work}
\label{sec:conclusion}
In this study, we evaluated and benchmarked the three state-of-the-art Vision-Language Models (VLMs) (Claude, Gemini, and GPT-4) and two traditional OCR models (RapidOCR and EasyOCR) on a custom OCR dataset containing 1,477 annotated frames. This dataset, created using VideoDB's \cite{videodb} infrastructure, will be publicly available through VideoDB along with our code. Our analysis showed that these models deliver strong performance, especially regarding average accuracy,  and outperform traditional computer vision models on dynamic video data. However, further work is needed to improve the robustness of these models, particularly against variations in video quality, font styles, and complex backgrounds. As the generalization capabilities of VLMs continue to improve, they are expected to become significant competitors, potentially replacing traditional methods in the near future. The growing sophistication of VLMs suggests that they may soon be capable of handling a broader range of tasks, leading to more versatile AI systems.


\noindent For future work, expanding the dataset by incorporating more diverse videos would provide a broader scope for evaluating the models. Additionally, fine-tuning VLMs on the proposed dataset could improve their adaptability and performance. Another potential direction is evaluating the effect of prompt variations on VLM performance, which could provide valuable insights for optimizing their responses. This research can be extended to other tasks where traditional computer vision models are typically employed, such as object detection, segmentation, and activity recognition, allowing us to assess whether VLMs can replace or complement these methods in real-world applications.

\bibliographystyle{plainurl} 
\bibliography{references}

\begin{thebibliography}{10}

\bibitem{anthropic2024claude}
{Anthropic}.
\newblock {Claude 3.5 Sonnet}: Advancements in multimodal {AI}, 2024.
\newblock URL: \url{https://www.anthropic.com/news/claude-3-5-sonnet}.

\bibitem{baek2019character}
Youngmin Baek, Bado Lee, Dongyoon Han, Sangdoo Yun, and Hwalsuk Lee.
\newblock Character region awareness for text detection.
\newblock In {\em Proceedings of the IEEE/CVF conference on computer vision and pattern recognition}, pages 9365--9374, 2019.

\bibitem{de2025video}
Ulindu De~Silva, Leon Fernando, Kalinga Bandara, and Rashmika Nawaratne.
\newblock Video summarisation with incident and context information using generative ai.
\newblock {\em arXiv preprint arXiv:2501.04764}, 2025.

\bibitem{google2024gemini}
{Google DeepMind}.
\newblock {Gemini 1.5 Pro}: Pushing the boundaries of multimodal learning, 2024.
\newblock URL: \url{https://deepmind.google/technologies/gemini/pro/}.

\bibitem{graves2012connectionist}
Alex Graves and Alex Graves.
\newblock Connectionist temporal classification.
\newblock {\em Supervised sequence labelling with recurrent neural networks}, pages 61--93, 2012.

\bibitem{EasyOCR2024}
{JaidedAI}.
\newblock {EasyOCR}: Ready-to-use {OCR} with 80+ supported languages, 2024.
\newblock URL: \url{https://github.com/JaidedAI/EasyOCR}.

\bibitem{lee2024vhelm}
Tony Lee, Haoqin Tu, Chi~Heem Wong, Wenhao Zheng, Yiyang Zhou, Yifan Mai, Josselin~Somerville Roberts, Michihiro Yasunaga, Huaxiu Yao, Cihang Xie, et~al.
\newblock Vhelm: A holistic evaluation of vision language models.
\newblock {\em arXiv preprint arXiv:2410.07112}, 2024.

\bibitem{lei2018tvqa}
Jie Lei, Licheng Yu, Mohit Bansal, and Tamara~L Berg.
\newblock Tvqa: Localized, compositional video question answering.
\newblock {\em arXiv preprint arXiv:1809.01696}, 2018.

\bibitem{openai2024gpt4}
{OpenAI}.
\newblock {GPT-4o}: Omni-modal language model, 2024.
\newblock URL: \url{https://openai.com/index/hello-gpt-4o/}.

\bibitem{paddleocr2023}
{PaddlePaddle} Team.
\newblock {PaddleOCR}: An {OCR} toolset based on {PaddlePaddle}, 2023.
\newblock URL: \url{https://github.com/PaddlePaddle/PaddleOCR}.

\bibitem{RapidOCR2021}
{RapidAI} Team.
\newblock {RapidOCR}: A lightweight {OCR} framework, 2021.
\newblock Open-source {OCR} solution.
\newblock URL: \url{https://github.com/RapidAI/RapidOCR}.

\bibitem{videodb}
{VideoDB}.
\newblock {VideoDB}: Video infrastructure for the {AI} first world, 2024.
\newblock A modern video processing and analysis platform.
\newblock URL: \url{https://videodb.io/}.

\bibitem{xu2016msr}
Jun Xu, Tao Mei, Ting Yao, and Yong Rui.
\newblock Msr-vtt: A large video description dataset for bridging video and language.
\newblock In {\em Proceedings of the IEEE conference on computer vision and pattern recognition}, pages 5288--5296, 2016.

\end{thebibliography}



\newpage
\appendix
\section{Supplementary Material}
\label{sec:supplementary_material}
\subsection{Dataset Examples}
\label{dataset_examples}
This includes few example frames from the custom dataset along with their annotation.

\begin{figure}[htbp]
    \centering
    \includegraphics[width=0.7\textwidth]{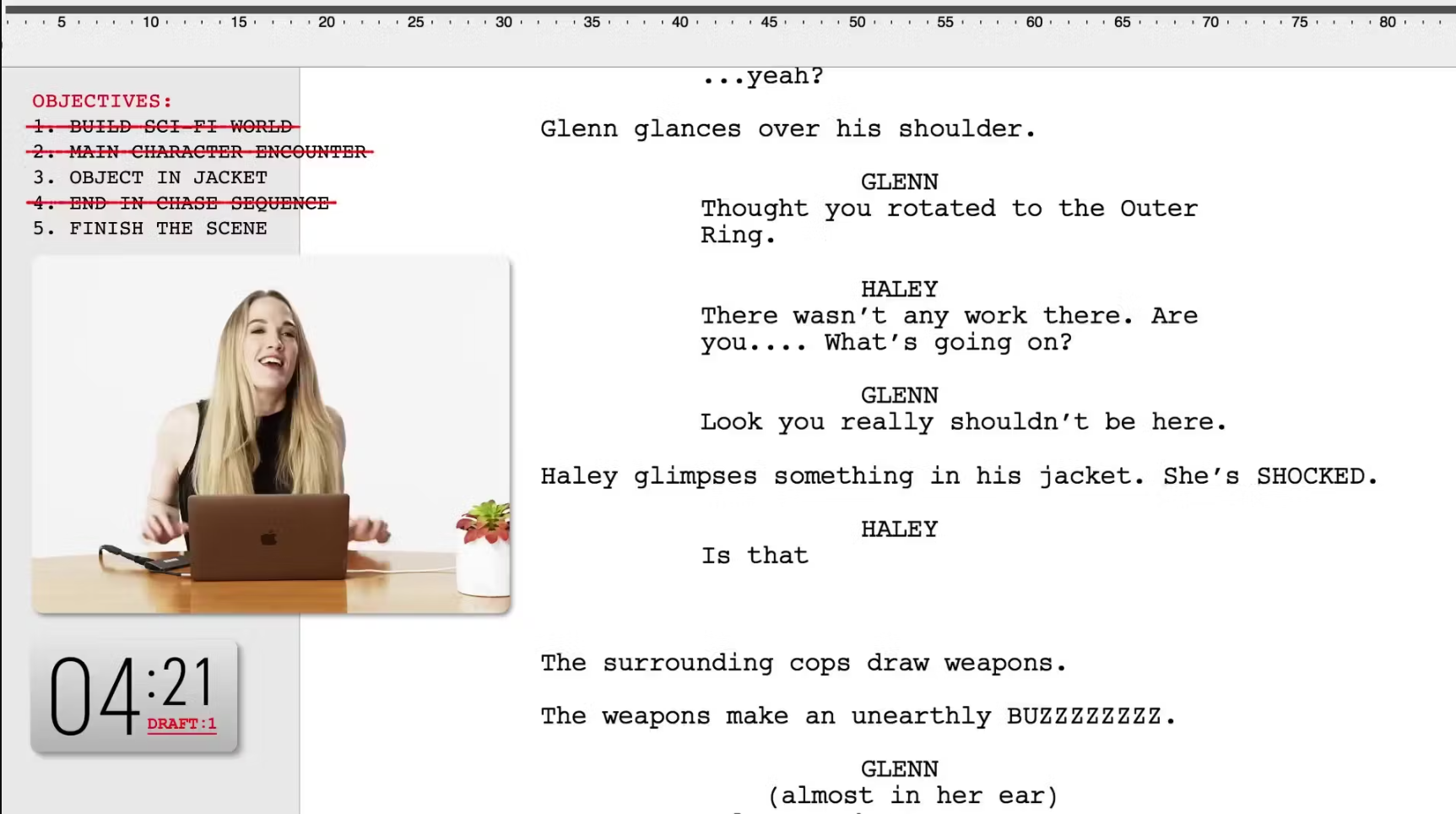}  
    \caption{Ground truth annotation: \textit{\textcolor{ForestGreen}{5 10 15 20 25 30 35 40 45 50 55 60 65 70 75 80 OBJECTIVES: 1. BUILD SCI FI WORLD 2. MAIN CHARACTER ENCOUNTER 3. OBJECT IN JACKET 4. END IN CHASE SEQUENCE 5. FINISH THE SCENE 04:21 DRAFT:1 ...yeah? Glenn glances over his shoulder. GLENN Thought you rotated to the Outer Ring. HALEY There wasn't any work there. Are you.... What's going on? GLENN Look you really shouldn't be here. Haley glimpses something in his jacket. She's SHOCKED. HALEY Is that The surrounding cops draw weapons. The weapons make an unearthly BUZZZZZZZZZ. GLENN (almost in her ear)}}}
    \label{fig:ocr_1}
\end{figure}

\begin{figure}[htbp]
    \centering
    \includegraphics[width=0.7\textwidth]{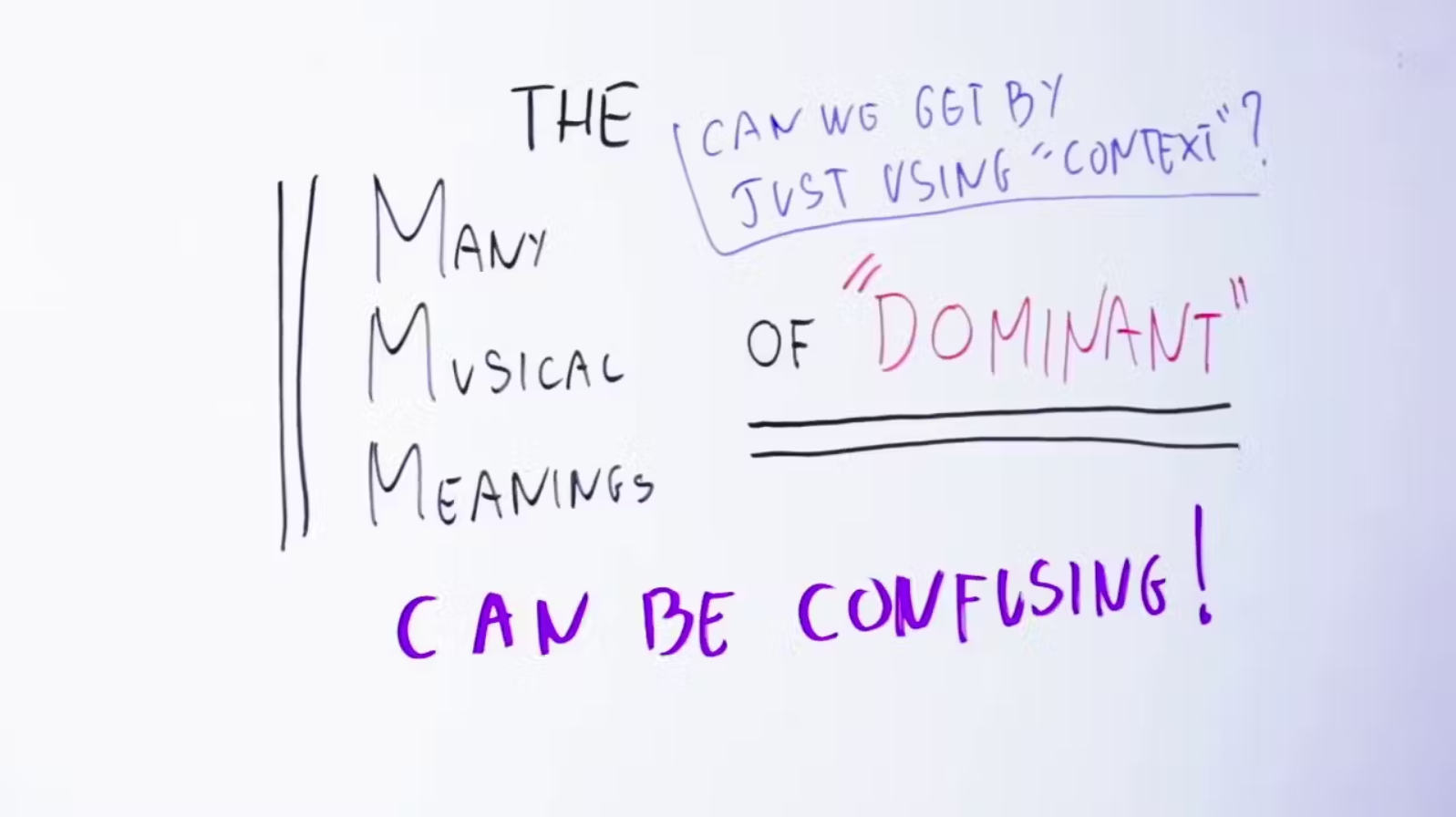}  
    \caption{Ground truth annotation: \textit{\textcolor{ForestGreen}{THE Many Musical Meanings CAN WE GET BY JUST USING "CONTEXT"? OF "DOMINANT" CAN BE CONFUSING!}}}
    \label{fig:ocr_2}
\end{figure}

\begin{figure}[htbp]
    \centering
    \includegraphics[width=0.7\textwidth]{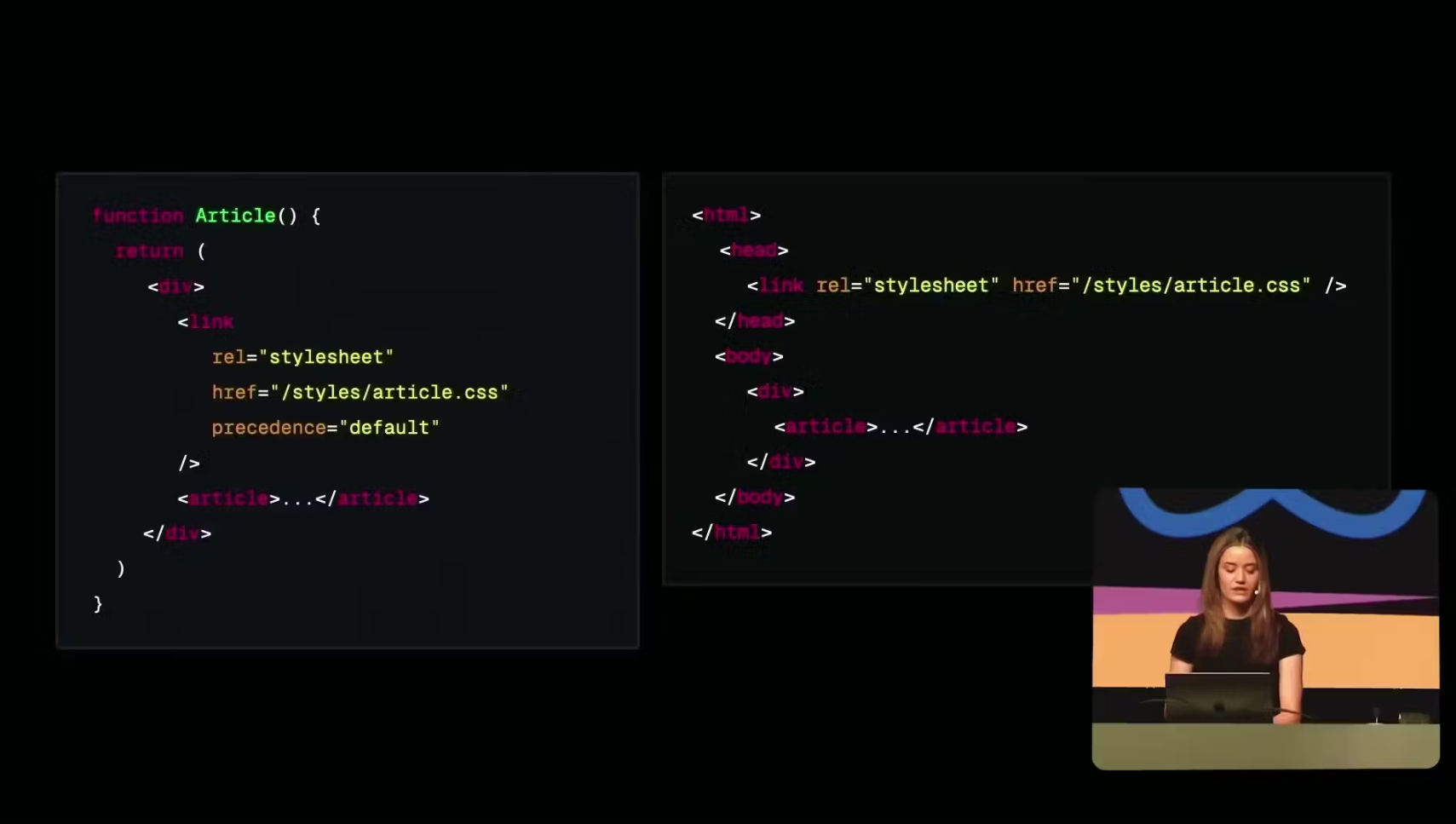}  
    \caption{Ground truth annotation: \textit{\textcolor{ForestGreen}{function Article() { return ( \textless div \textgreater \textless link rel="stylesheet" href="/styles/article.css" precedence="default" \textbackslash  \textgreater \textless article \textgreater...\textless \textbackslash article \textgreater \textless \textbackslash div \textgreater ) } \textless html \textgreater \textless head \textgreater \textless link rel="stylesheet" href="/styles/article.css" / \textgreater \textless /head \textgreater \textless body \textgreater \textless div \textgreater \textless article \textgreater...\textless /article \textgreater \textless /div \textgreater \textless /body \textgreater \textless /html \textgreater}}}
    \label{fig:ocr_3}
\end{figure}

\newpage \subsection{Additional Results}
\label{additional_results}
This section contains detailed results not included in the main paper.

\begin{figure}[htbp]
    \centering
    \includegraphics[width=0.7\textwidth]{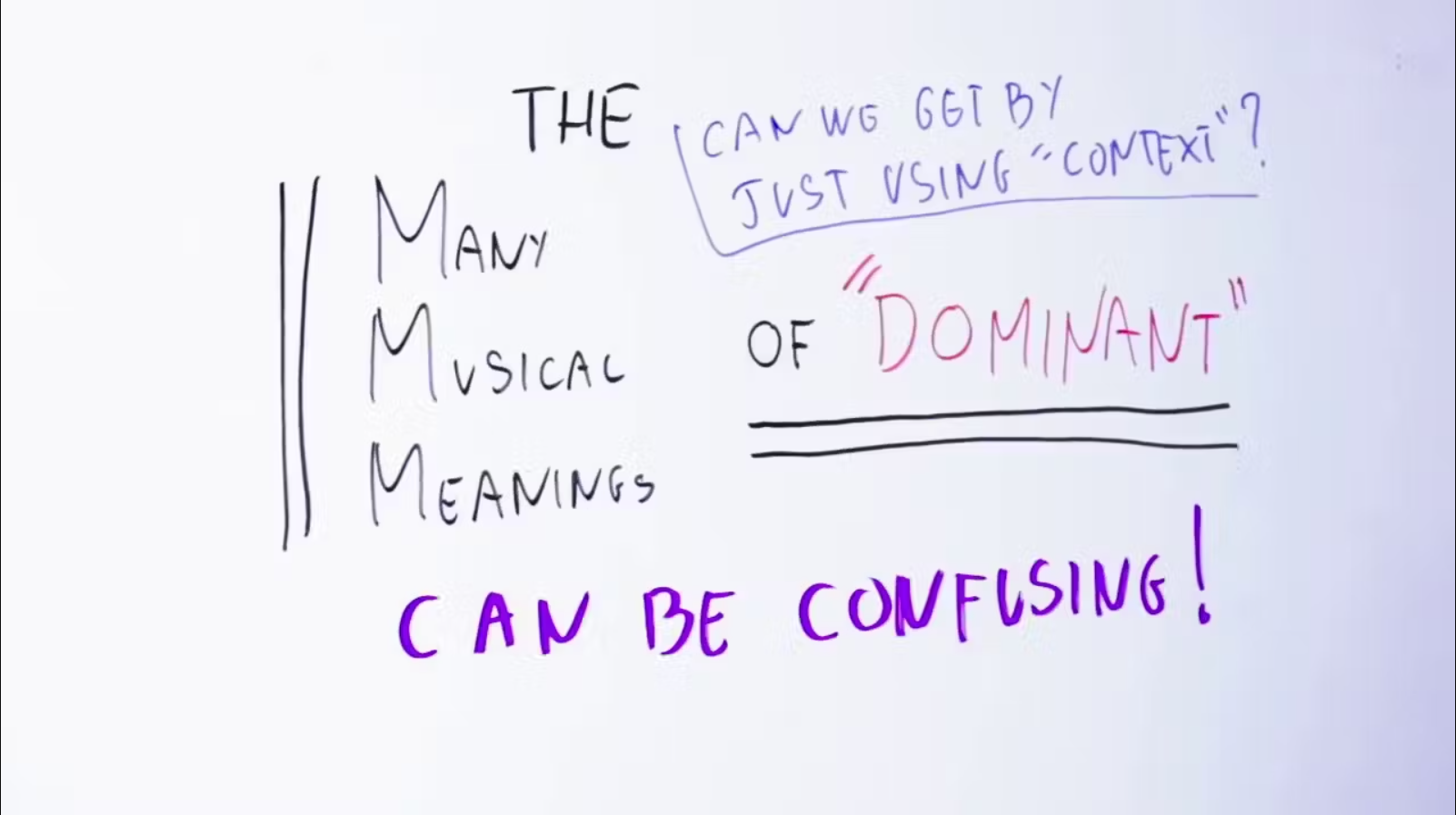}  
    \caption{ \textbf{Ground Truth:} \textcolor{ForestGreen}{\textit{THE Many Musical Meanings CAN WE GET BY JUST USING "CONTEXT"? OF "DOMINANT" CAN BE CONFUSING!}}}
    \label{fig:ocr_4}
\end{figure}

\noindent \textbf{Claude Output:} \\
\textcolor{black}{\textit{THE can we get by just using "context"? Many Musical Meanings of "DOMINANT" can be confusing!}}\newline

\noindent \textbf{Gemini Output:} \\
\textcolor{black}{\textit{THE CAN WE GET BY "CONTEXT" ? JUST USING "CONTEXT" MANY MUSICAL MEANINGS OF "DOMINANT" CAN BE CONFUSING !}}\newline

\noindent \textbf{GPT-4 Output:} \\
\textcolor{black}{\textit{THE Many Musical Meanings CAN WE GET BY JUST USING "CONTEXT"? OF "DOMINANT" CAN BE CONFUSING!}}\newline

\noindent \textbf{RapidOCR Output:} \\
\textcolor{black}{\textit{CANWG GGTBY X,SI5 ANY DOMINANT M OF MEAMINGS CAN BE CONFLSING!}}\newline

\noindent \textbf{EasyOCR Output:} \\
\textcolor{black}{\textit{THE WG 661 By '7 Covtxi" MAwy Mvsicac OF 'DoMINiUt' MeANigs C AV DE CovFlsing CAv vSing JvSi (}}\newline

\noindent In figure \ref{fig:ocr_4}, Claude's output captures most of the content but introduces a significant rearrangement in sentence structure. Specifically, it shifts the position of "can we get by just using "context"?" to the beginning of the sentence and adjusts the flow, altering the natural sequence of ideas in the ground truth. Gemini's output demonstrates a notable deviation, repeating the phrase "CAN WE GET BY "CONTEXT"?" unnecessarily and omitting part of the original meaning. Furthermore, it fails to preserve the original sentence's formatting and clarity. In contrast, GPT-4's output is nearly identical to the ground truth, accurately maintaining the text, capitalization, and punctuation, making it the closest match to the original annotation. The traditional computer vision models, however, demonstrated significant shortcomings,
failing to recognize even simple handwritten text. \\

\begin{figure}[htbp]
    \centering
    \includegraphics[width=0.7\textwidth]{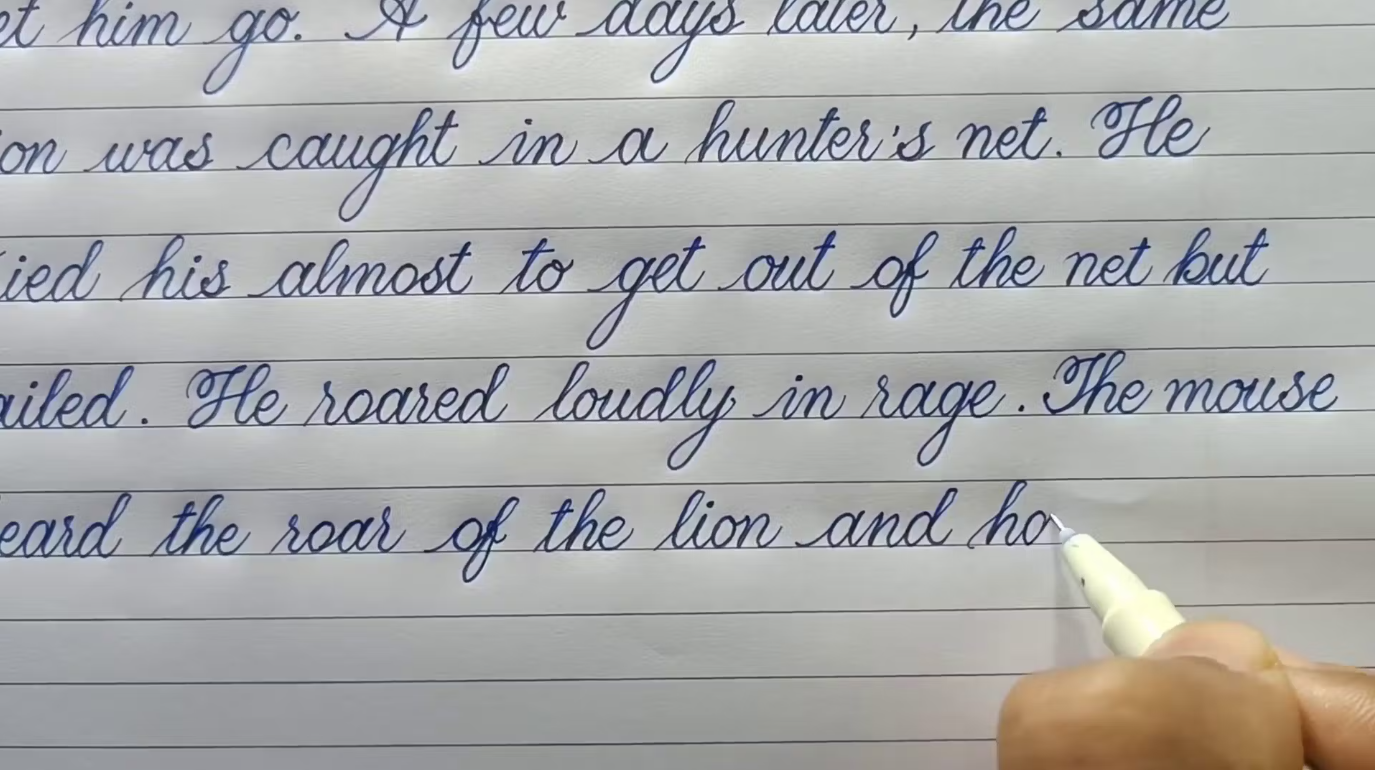}  
    \caption{ \textbf{Ground Truth:} \textcolor{ForestGreen}{\textit{t him go. A few days later, the same on was caught in a hunter's net. He ied his almost to get out of the net but iled. He roared loudly in rage. The mouse eard the roar of the lion and ho}}}
    \label{fig:ocr_7}
\end{figure}

\noindent \textbf{Claude Output:} \\
\textcolor{black}{\textit{"A lion was caught in a hunter's net. He tried his almost to get out of the net but failed. He roared loudly in rage. The mouse heard the roar of the lion and he"}}\newline

\noindent \textbf{Gemini Output:} \\
\textcolor{black}{\textit{let him go A few days later the Same lion was caught in a hunter's net He tied his almost to get out of the net but failed He roared loudly in rage The mouse heard the roar of the lion and ho}}\newline

\noindent \textbf{GPT-4 Output:} \\
\textcolor{black}{\textit{let him go. A few days later, the same lion was caught in a hunter's net. He tried his almost to get out of the net but failed. He roared loudly in rage. The mouse heard the roar of the lion and he}}\newline

\noindent \textbf{RapidOCR Output:} \\
\textcolor{black}{\textit{go eu omx hunter's net.gle tina onuascaught ied ailed.gle roaredloudly mouse}}\newline

\noindent \textbf{EasyOCR Output:} \\
\textcolor{black}{\textit{I him Go XAe1AUOLoJlkI,knes OMIE; OU MHs inAU huntehs nebd4es ied his alost to oubf the nebdbut lede 9le hodied loudhp Znshag Ihemose eaid Zhes hoab the lion andzho caught ez}}\newline

\noindent In figure \ref{fig:ocr_7}, Claude's output deviates significantly from the ground truth, omitting "let him go" and inaccurately reconstructing phrases like "tried his almost." While fluent, it lacks precision. Gemini preserves more of the original structure, including "let him go," but introduces errors such as "tied his almost" and lacks punctuation, affecting clarity. GPT-4 provides the most accurate reconstruction, retaining key phrases, punctuation, and structure, but slightly truncates the final word. Overall, GPT-4 outperforms the others, followed by Gemini, while Claude shows the greatest deviations. On the other hand, the traditional computer vision models fails to produce coherent text, outputting gibberish instead.

\begin{figure}[htbp]
    \centering
    \includegraphics[height=0.3\textheight]{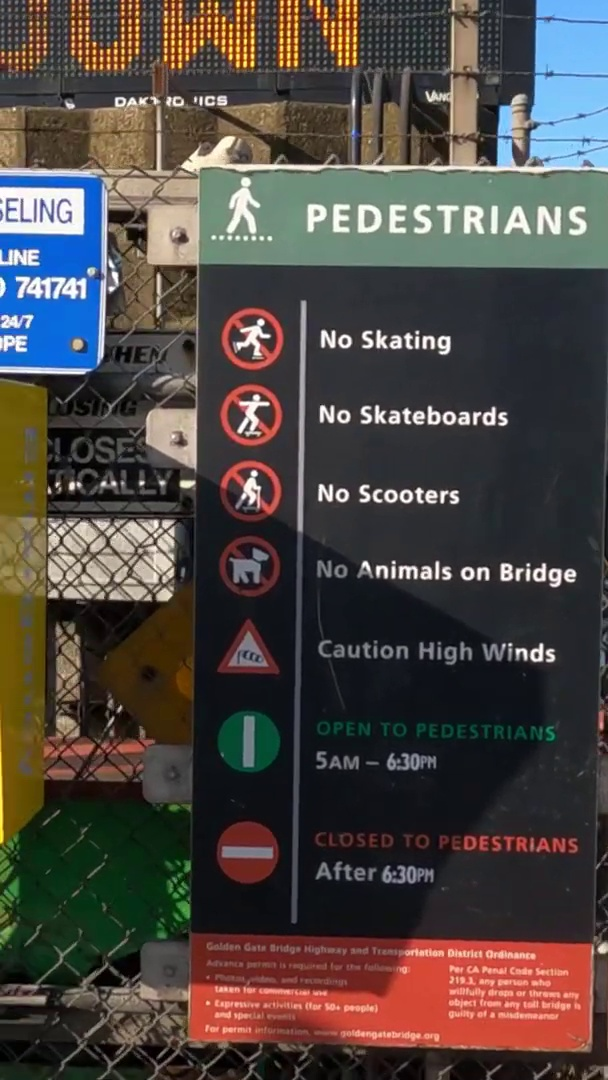}  
    \caption{\textbf{Ground Truth:} \textcolor{ForestGreen}{\textit{PEDESTRIANS No Skating No Skateboards No Scooters No Animals on Bridge Caution High Winds OPEN TO PEDESTRIANS 5AM - 6:30PM CLOSED TO PEDESTRIANS After 6:30PM Golden Gate Bridge Highway and Transportation District Ordinance Per CA Penal Code Section 219.3, any person(s) who willfully drops or throws any object or missile from any toll bridge is guilty of a misdemeanor. For permits information visit www.goldengatebridge.org}}}
    \label{fig:ocr_9}
\end{figure}

\noindent \textbf{Claude Output:} \\
{\textit{VEHIC PEDESTRIANS No Skating No Skateboards No Scooters No Animals on Bridge Caution High Winds OPEN TO PEDESTRIANS 5am - 6:30pm CLOSED TO PEDESTRIANS After 6:30pm Golden Gate Bridge Highway and Transportation District Activities permits is required for film permits Section No person shall carry or possess any alcoholic beverages for any reason (Vict Sec 21399.5) failure to comply may result in a misdemeanor}}\newline

\noindent \textbf{Gemini Output:} \\
\textcolor{black}{\textit{SELING LINE 741741 24/7 PEDESTRIANS No Skating No Skateboards No Scooters No Animals on Bridge Caution High Winds OPEN TO PEDESTRIANS 5AM \u2013 6:30PM CLOSED TO PEDESTRIANS After 6:30PM Golden Gate Bridge Highway and Transportation District Ordinance Advance petesis is repared for the following Pes, and recordings taken for commercial ne Expressive activities (for 50+ people) and special eve For permit information goldengatebridge.org Per CA Penal Code Section 219.3, any person who willfully draps or threws any object from any toll bridge is guilty of a misdemeanor}}\newline

\noindent \textbf{GPT-4 Output:} \\
\textit{DOWN PEDESTRIANS No Skating No Skateboards No Scooters No Animals on Bridge Caution High Winds OPEN TO PEDESTRIANS 5AM 630PM CLOSED TO PEDESTRIANS After 630PM Golden Gate Bridge Highway and Transportation District Ordinance Advance permit is required for the following Per CA Penal Code Section 219.3 any person who filming and proceedings willfully drops or throws any taken for commercial use object from any toll bridge is Expressive activities for 50 people guilty of a misdemeanor For permit information goldengatebridge org}\newline

\noindent \textbf{RapidOCR Output:} \\
\textcolor{black}{\textit{DAKRDICS ELING PEDESTRIANS LINE 741741 24/7 No Skating PE HEN No Skateboards LOSE No Scooters No Animals on Bridge Caution High Winds OPEN TO PEDESTRIANS 5AM-6:30PH CLOSED TO PEDESTRIANS After6:30PM oration Distriet Grdinance Per CA Pensl Cdle Seetinn 219.Lany pertonmho objeci froany tltrdgi guilty ofamusdemeanor}}\newline

\noindent \textbf{EasyOCR Output:} \\
\textcolor{black}{\textit{DOWN PEDESTRIANS No Skating No Skateboards No Scooters No Animals on Bridge Caution High Winds OPEN TO PEDESTRIANS 5AM 630PM CLOSED TO PEDESTRIANS After 630PM Golden Gate Bridge Highway and Transportation District Ordinance Advance permit is required for the following Per CA Penal Code Section 219.3 any person who filming and proceedings willfully drops or throws any taken for commercial use object from any toll bridge is Expressive activities for 50 people guilty of a misdemeanor For permit information goldengatebridge org}}\newline

\noindent Lastly, in figure \ref{fig:ocr_9},  
Claude performed best among the large language models, accurately capturing core information about pedestrian access and the prohibition of throwing objects, but hallucinating additional rules about filming and alcohol. Gemini produced a significantly less accurate transcription, riddled with misspellings and wrong details. Both the traditional computer vision OCR models struggled considerably, misreading words and generating nonsensical outputs due to the sign's format. \\
\end{document}